\documentclass[10pt]{article} 
\usepackage[preprint]{tmlr}

\usepackage{amsmath,amsfonts,bm}

\def\eqref#1{equation~\ref{#1}}

\def\1{\bm{1}}

\DeclareMathAlphabet{\mathsfit}{\encodingdefault}{\sfdefault}{m}{sl}
\SetMathAlphabet{\mathsfit}{bold}{\encodingdefault}{\sfdefault}{bx}{n}

\usepackage{hyperref}
\usepackage{url}
\usepackage{booktabs}
\usepackage{graphicx}
\usepackage{multirow}
\usepackage{enumitem}
\title{GRIP: Feedback-Guided Prompt Retrieval for Large \\ Multimodal Models}

\author{
\begin{tabular}{c}
Garvita Allabadi\textsuperscript{1},
Matteo Sodano\textsuperscript{2},
Roberto Estevão\textsuperscript{3}, \\
Yuxiong Wang\textsuperscript{1},
Vikram Adve\textsuperscript{1},
Emre K{\i}c{\i}man\textsuperscript{3},
Ranveer Chandra\textsuperscript{3} \\
\textsuperscript{1}University of Illinois Urbana Champaign \quad
\textsuperscript{2}University of Bonn \quad
\textsuperscript{3}Microsoft
\end{tabular}
}

\def\month{MM}  
\def\year{YYYY} 
\def\openreview{\url{https://openreview.net/forum?id=XXXX}} 

\begin{document}

\maketitle

\begingroup
\renewcommand\thefootnote{}%
\footnotetext{Correspondence: garvita4@illinois.edu}%
\endgroup

\begin{abstract}
In-Context Learning (ICL) has become a powerful mechanism for adapting Large Language Models (LLMs) to new tasks without fine-tuning. Extending this concept to Large Multimodal Models (LMMs), Multimodal In-Context Learning (M-ICL) relies on retrieving relevant examples, such as images, captions, or question-answer pairs, to guide predictions across tasks like classification, captioning, and visual question answering (VQA). Most existing approaches select in-context examples based on feature-space similarity, assuming that semantically similar samples provide the most useful context. However, our systematic analysis reveals that this assumption does not always hold: visually similar examples are not necessarily those that most effectively enhance in-context learning performance.  

To address this, we propose the Guided Retrieval of In-context Prompts (GRIP), a learnable vision-only retrieval framework that leverages feedback from LMMs to identify examples that truly improve model predictions. GRIP learns to distinguish beneficial from detrimental in-context examples through contrastive training, refining retrieval beyond pure similarity. Across three multimodal tasks, namely classification, captioning, and VQA, GRIP improves consistently over similarity-based retrieval on Qwen2.5-VL-7B, with its strongest gains in classification on Idefics2-8B. Moreover, we demonstrate that retrievers trained with feedback from one open LMM can be transferred to other models without retraining, including closed-source GPT-4o and Gemini, enabling scalable and cost-efficient deployment of M-ICL. Code will be published upon acceptance.

\end{abstract}

\section{Introduction}

Large pre-trained language models (LLMs) demonstrated that powerful, general-purpose models can be adapted to new tasks not by parameter updates but by carefully chosen context, given by a few examples presented in the prompt. This capability, called In-Context Learning (ICL) and first introduced by~\cite{brown2020language}, is attractive because it enables rapid task adaptation without fine-tuning, preserves a single shared model, and avoids the data, compute, and maintenance costs associated with per-task training. As research moved beyond text, the same idea naturally extends to large multimodal models (LMMs) (\cite{alayrac2022flamingo, sung2022vl, zhao2023mmicl}): instead of only text examples, a prompt can contain image–text pairs that demonstrate the task before the actual query. Multimodal ICL (M-ICL), therefore, promises the same flexibility for vision-language tasks such as classification, captioning, and visual question answering (VQA), allowing a single model to perform many tasks simply by changing the examples shown at inference time.

Despite its conceptual simplicity, the main practical question of M-ICL regards what should be shown in the prompt. The choice of in-context examples has a significant influence on model performance. A standard strategy, nowadays, consists of selecting in-context examples based on their proximity to the test sample in the embedding space of the LLM. This strategy, often called prompt retrieval, aims to identify and curate the most informative samples from the training set to serve as in-context examples for the query. 

In M-ICL, however, the effectiveness of a retrieved example is not necessarily determined by its proximity in a visual feature space. While semantically similar images often provide relevant context, similarity alone is not a sufficient criterion to guarantee improved model performance. LMMs integrate visual and textual information in a complex manner, and the usefulness of an example depends on how well it complements the model’s internal representations rather than on surface-level resemblance. Thus, retrieval strategies that rely exclusively on nearest neighbors in pre-trained embedding spaces are inherently limited.

Our key conceptual shift is to treat retrieval for M-ICL as a problem of utility alignment: examples should be selected not because they are perceptually similar, but because they demonstrably improve the LMM’s behavior on the task. This suggests a natural source of supervision, which is the LMM itself. The resulting signal is intrinsically task-aware, and it can be computed using task-appropriate scoring functions, is interpretable (positive, neutral, or negative contribution), and is readily available whenever we can query the LMM on labeled training samples.

Building on this principle, we propose a concrete, operational recipe. First, constrain the candidate pool with a cheap, off-the-shelf visual encoder so that the search space is tractable and contains semantically plausible examples. Second, convert the LMM’s behavior into supervision by evaluating the model with and without each candidate neighbor and assigning labels according to a task-appropriate score change. Third, train a compact vision-only retriever using a supervised contrastive objective that pulls beneficial examples closer to their queries and pushes harmful ones away; during training we emphasize hard negatives (neighbors that are perceptually close but unhelpful) to reshape the representation precisely where similarity-based methods fail. At inference time the learned retriever performs efficient nearest-neighbor search in the new embedding space; the retrieved examples are then supplied as the in-context prompt to the LMM.

In this article, we propose the \emph{Guided Retrieval of In-context Prompts} (GRIP), a feedback-guided retriever that (1) uses the target LMM’s own outputs to generate labeled positives and negatives for candidate neighbors, (2) trains a contrastive, vision-only retriever that aligns embedding geometry with in-context utility rather than raw similarity, and (3) is designed for practical deployment, as it uses an initial pre-filtering step to limit labeling cost, leverages hard-negative mining, and supports transfer when the target model is expensive or closed-source by training on surrogate models. We validate this approach across several common vision–language tasks and show that a retriever learned from model feedback selects examples that more reliably improve downstream performance than standard similarity heuristics.

In sum, our contributions are:
\begin{itemize}[leftmargin=*, itemsep=0.15em, topsep=0.25em, parsep=0pt, partopsep=0pt]
    \item We provide a systematic quantitative and qualitative analysis of the similarity-based retrieval paradigm in multimodal in-context learning. Our study shows that semantically similar neighbors are not reliably the most beneficial examples for large multimodal models. This challenges a widely adopted assumption in the field and motivates the development of retrieval strategies that go beyond pure similarity.
    \item We introduce a simple, vision-only feedback-guided retriever for multimodal ICL, called GRIP, that improves ICL performance using feedback from LMMs. Unlike prior similarity-driven or unsupervised heuristics, GRIP learns to distinguish beneficial from detrimental in-context examples, improving consistently over similarity-based retrieval on Qwen2.5-VL-7B, with its strongest Idefics2-8B gains in classification.
    \item We show that retrievers trained with feedback from one LMM can transfer to another, including closed-source API-based systems like GPT-4o and Gemini. This enables scalable, cost-efficient performance improvements without the need for retraining or expensive per-model adaptation, pointing toward practical deployment of M-ICL at scale.    
\end{itemize}

\section{Basics for In-Context Learning}

\subsection{Multimodal In-Context Learning}

In-context learning (ICL) has emerged as one of the most influential ideas in the development of large language models (LLMs). Instead of adapting models through parameter updates, ICL allows models to perform new tasks by conditioning on a small number of input–output demonstrations included directly in the prompt. This paradigm provides flexibility, since the same model can be applied to a wide range of tasks by simply varying the examples provided at inference time, without requiring additional training or fine-tuning.

Building on this success, the idea has been extended to large multimodal models (LMMs), which are trained to jointly process and reason over data from multiple modalities, most prominently images and text. \emph{Multimodal in-context learning} (M-ICL) enables such models to generalize to novel multimodal tasks by leveraging a handful of demonstrations that combine both textual and visual information. Tasks that benefit from this paradigm include image classification, captioning, and visual question answering (VQA), where models can adapt to unseen instructions or domains solely from in-context examples.

Formally, let $D = \{(x_i, y_i)\}_{i=1}^{N}$ denote a dataset consisting of $N$ multimodal input–output pairs, where $x_i$ represents a multimodal input (e.g., an image with an associated text query) and $y_i$ is the corresponding output (e.g., a caption or an answer). Given a query $x_q$, M-ICL involves selecting a subset $S \subset D$ of in-context examples to provide as part of the prompt for the model $M$. The model then predicts
\begin{equation} 
M(S, x_q) \to \hat{y}_q,
\end{equation}
where $\hat{y}_q$ is the model’s prediction for the query, conditioned on the examples $S$.

As a concrete example, consider a VQA setting. A query $x_q$ consists of an image–question pair, $x_q = (\text{image}_q, \text{question}_q)$. The model is provided with a small set of $k$ demonstrations, $\mathcal{S} = \{(\text{image}_j, \text{question}_j, \text{answer}_j)\}_{j=1}^k$, each encoding the input–output structure of the task. Conditioned on this set, the LMM generates an answer $\hat{y}_q$ for the query. The effectiveness of this process hinges critically on the choice of $\mathcal{S}$, as the selected examples shape the inductive bias the model applies when solving the task.

\begin{table}
    \centering
    \caption{Average rank of positive examples among top-16 retrieved neighbors across datasets. Lower values indicate that beneficial examples are ranked closer to the top.}
    \vspace{0.2cm}
    \label{tab:avg-rank}
    \begin{tabular}{ccccccc}
    \toprule
    Model & UC Merced & Pets & DTD & SEED & ScienceVQA & COCO Caption \\
    \midrule
    Qwen2.5-VL-7B & 8.19 & 8.04 & 7.75 & 8.48 & 8.45 & 7.99 \\
    Idefics2-8B             & 8.15 & 8.17 & 7.15 & 8.47 & 8.44 & 8.01 \\
    \bottomrule
    \end{tabular}
\end{table}

\subsection{Analyzing Visual Prompt Retrieval for Large Multimodal Models} 

The central challenge in M-ICL is thus not only the design of the model but also the selection of appropriate in-context examples. Effective retrieval directly determines the quality of the prompt, and thereby the generalization ability of the LMM to the target query. Poorly chosen examples may confuse the model, introduce irrelevant context, or bias predictions, while carefully chosen examples can substantially improve task performance. In practice, retrieval is often framed as the problem of identifying a small set of examples from a large training pool that are most informative for the given query.

\begin{figure*}[!ht]
    \centering
    \includegraphics[width=\linewidth]{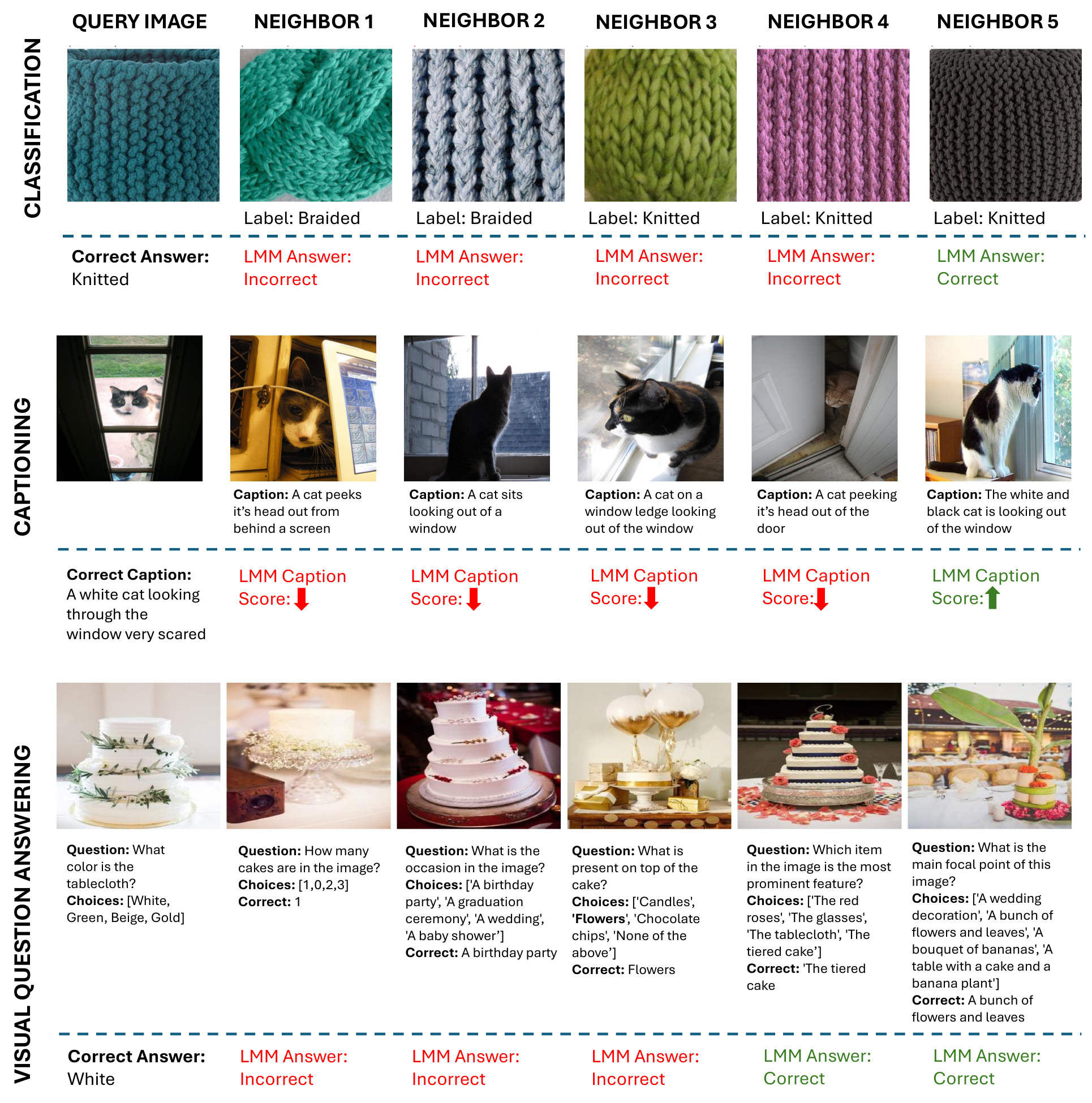}
    \caption{Qualitative analysis of similarity-based retrieval in multimodal in-context learning. Each row shows a query (left) along with its top retrieved neighbors. Even when neighbors share the correct label or depict semantically related content, the LMM often fails to predict the correct output. For classification (row 1), retrieved images include visually similar and correctly labeled samples, yet the LMM sometimes produces the wrong class. In captioning (row 2), neighbors describe scenes that are visually similar but do not provide the nuanced contextual cues required for accurate captions. In visual question answering (row 3), semantically related questions and answers are retrieved, but the model’s reasoning remains incorrect for several neighbors. These examples highlight a key limitation of the similarity paradigm: semantic proximity in embedding space does not reliably translate into in-context usefulness for LMMs. This motivates the need for retrieval strategies that explicitly consider the \emph{utility} of examples, as we develop in the next section.
    }
    \label{fig:lmm-analysis}
\end{figure*}

Existing approaches to example selection in image-based tasks have largely relied on unsupervised similarity measures. A common strategy, employed by Retrieval-based In-Context Example Selection (RICES) and proposed by \cite{yang2022empirical}, is to embed both the query and the candidate examples in a pre-trained feature space and to retrieve those nearest in distance. When applied to multimodal models, this means that examples for tasks such as image captioning, classification, or VQA are typically chosen based on the visual similarity of the images (using encoders such as CLIP or self-supervised vision transformers), sometimes complemented with textual similarity. The underlying assumption is that semantically related inputs provide useful context for the model, and that proximity in the embedding space is a sufficient proxy for relevance in in-context learning.

While this strategy has shown promise, it implicitly assumes that semantic similarity aligns with in-context utility, i.e., that the most visually or textually similar examples are also those that most improve the model’s prediction when provided in the prompt. Whether this assumption holds in practice, and how to design retrieval strategies that better capture what truly benefits multimodal ICL, is the core problem we investigate in this work.
\section{Challenging the Similarity Paradigm}

As outlined in the previous section, most existing approaches to prompt retrieval, such as RICES (\cite{yang2022empirical}), operate under a simple assumption: semantically similar examples provide useful context for the query. While intuitive, this assumption implicitly equates similarity in a pre-trained embedding space with usefulness for in-context learning. In practice, however, large multimodal models (LMMs) do not necessarily exploit visual similarity in such a direct way. The value of an example depends not only on its resemblance to the query but also on how it interacts with the model’s internal reasoning and the demands of the specific task. In the following, we examine to what extent the similarity paradigm aligns with actual in-context utility.

Our analysis investigates how retrieved neighbors affect model predictions. For each training example, we identify its top-$K$ most similar samples and evaluate the model’s output when each neighbor is included as an in-context example. By comparing predictions against ground truth, we categorize each neighbor as \emph{positive} if it improves performance or \emph{negative} if it does not. This procedure allows us to directly test whether the most similar neighbors are indeed the most beneficial for in-context learning. Note that we feed the neighbors to the model as in-context examples one by one, individually, rather than cumulatively.

If the similarity assumption were correct, the highest-ranked neighbors would consistently correspond to positives. However, Table~\ref{tab:avg-rank} shows that this is not the case. Across datasets, the average rank of positive neighbors is approximately 8 out of 16, indicating that helpful examples are scattered throughout the list rather than concentrated at the top. Thus, semantic proximity does not reliably predict in-context usefulness.

Qualitative examples further illustrate this misalignment (Figure~\ref{fig:lmm-analysis}). In classification, even correctly labeled neighbors may fail to improve the model’s predictions. In captioning, visually similar scenes often lack the contextual cues needed for accurate descriptions. In VQA, retrieved examples can share surface similarities with the query but still misguide the model’s reasoning. These cases highlight that relying solely on similarity overlooks the nuances of how LMMs integrate in-context information.

In sum, while similarity-based retrieval provides a convenient baseline, it is not sufficient for robust multimodal in-context learning. This motivates a shift toward retrieval strategies that are directly optimized for their contribution to model performance. In the next section, we present such an approach: a feedback-guided retriever that uses the model’s own outputs to identify and learn from effective in-context examples.

\section{Our Approach to In-Context Retrieval for LMMs}
\begin{figure*}[!t]
    \centering
    \includegraphics[width=\linewidth]{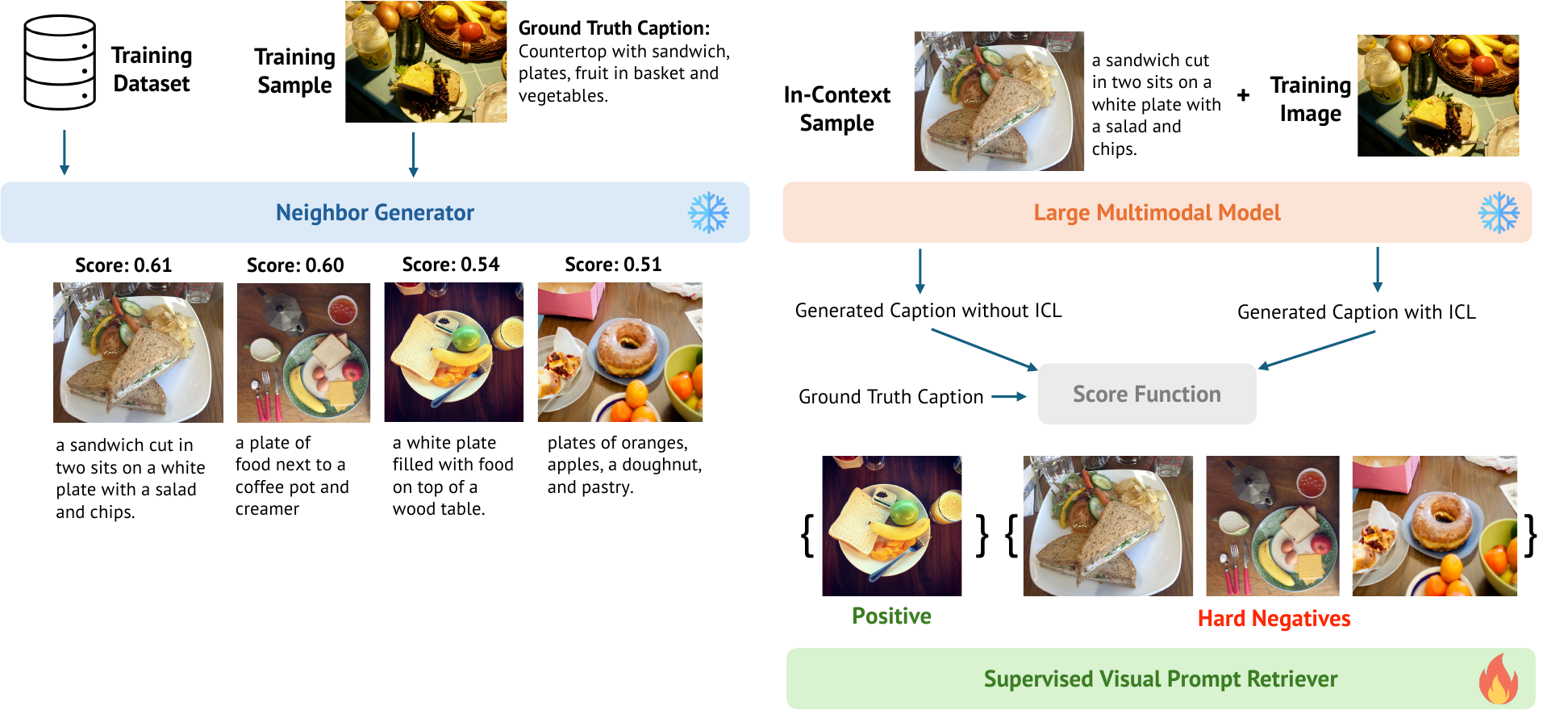}
    \caption{Overview of the proposed feedback-guided visual prompt retriever (GRIP). Given a training image, we first retrieve a set of visually similar candidate samples (\emph{neighbors}) from the training dataset using a pre-trained encoder. Each neighbor is then evaluated as an in-context example by a large multimodal model (LMM), which generates predictions both with and without the retrieved sample in context. The model’s responses are compared against the ground-truth label or caption through a task-specific scoring function to quantify the contribution of each neighbor. Samples that improve model performance are labeled as \emph{positive}, while those that degrade or do not influence performance are treated as \emph{hard negatives}. These labeled pairs are then used to train a supervised visual prompt retriever that learns to select the most beneficial in-context examples for future queries. This feedback-guided learning process allows GRIP to refine retrieval beyond visual similarity, leading to improved in-context learning performance across tasks and models.
}
    \label{fig:grip}
\end{figure*}

To overcome the limitations of similarity-driven retrieval, we propose the \emph{Guided Retrieval of In-context Prompts} (GRIP), a learnable retriever that selects in-context examples not based on raw feature similarity but on their demonstrated utility for the target model. The central idea is to let the model itself provide supervision: examples are judged by whether they improve or degrade the model’s predictions when included in the prompt. This transforms prompt retrieval from a heuristic nearest-neighbor search into a supervised learning problem directly aligned with in-context performance. An illustration of our approach is shown in Fig.~\ref{fig:grip}.

\subsection{GRIP: Our Novel Retrieval Strategy}

\paragraph{Data.} A core challenge in this setting is that no labeled dataset exists for the task of identifying beneficial in-context examples. To address this, we generate training labels automatically using feedback from the model. Specifically, for each training instance, we first retrieve the top $K$ nearest neighbors using a pre-trained visual encoder. This step provides a semantically plausible candidate pool while keeping the computational cost tractable.  

We then evaluate the target model’s prediction for the training instance in two conditions: (i) without any in-context examples and (ii) with each of the retrieved neighbors added individually as a prompt example. Since the ground-truth label of the training instance is available, we can score the effect of each neighbor by comparing the two predictions. Neighbors that improve the prediction are assigned as \emph{positives}, while those that fail to help or actively mislead the model are treated as \emph{hard negatives}. These are ``hard'' because they are close in the pre-trained feature space but unhelpful in practice, and thus crucial to separate during training. Through this procedure, every training instance yields supervision that reflects actual in-context utility rather than proxy similarity.  

\paragraph{Scoring for Classification.}
In the classification setting, each input corresponds to an image $I$ with an associated ground-truth label $G$. The retriever’s role is to identify which neighbors provide the most useful context for predicting the correct class of $I$. To this end, we evaluate the LMM both without and with a candidate neighbor $N$ as an in-context example. Let $P_0 = M(I)$ denote the prediction for $I$ in isolation, and $P_N = M(N, I)$ the prediction when $N$ is included in the prompt. A neighbor is labeled as \emph{positive} if its inclusion leads to the correct prediction, and \emph{negative} otherwise. Formally, the scoring function for classification can be expressed as

\begin{equation}
    s(N, I) =
    \begin{cases}
        1, & \text{if } P_N = G, \\
        0, & \text{otherwise}.
    \end{cases}
\end{equation}

This procedure provides binary supervision for training the retriever: positives highlight neighbors that guide the model toward the correct label, while negatives capture cases where similarity alone is insufficient to improve in-context performance.

\paragraph{Scoring for Captioning.}  
In the captioning task, each input $I$ is paired with a ground-truth caption $G$. The retriever must identify neighbors that improve the quality of the captions generated by the LMM when used as in-context examples. To measure this, we first obtain the model’s prediction without any neighbors, $P_0 = M(I)$, and then its prediction with a candidate neighbor $N$, $P_N = M(N, I)$. Since captions are open-ended and cannot be evaluated by exact matching, we rely on a language-model-based judge to score the similarity between a prediction and the ground truth. Specifically, we use the Llama-3.2-3B-Instruct model (\cite{grattafiori2024llama}) and, given a generated caption and ground truth caption, we prompt the model to evaluate the quality of captions by assigning scores based on relevance, fluency, coverage, conciseness, and overall similarity, but our method is not tailored to a specific LLM. The final score is an average of the above scores. Let $\text{LLM-Judge}(P, G)$ denote the score assigned to a caption $P$ with respect to the reference $G$. We compute two scores, $\text{Score}_1 = \text{LLM-Judge}(P_0, G)$ and $\text{Score}_2 = \text{LLM-Judge}(P_N, G)$. A neighbor is labeled as \emph{positive} if the judged score improves with its inclusion, and \emph{negative} otherwise:

\begin{equation}
    s(N, I) =
    \begin{cases}
        1, & \text{if } \text{LLM-Judge}(P_N, G) > \text{LLM-Judge}(P_0, G), \\
        0, & \text{otherwise}.
    \end{cases}
\end{equation}

This formulation allows the retriever to learn from relative improvements in caption quality, ensuring that positive neighbors are those that contribute additional context which enhances descriptive accuracy. 

\paragraph{Scoring for Visual Question Answering.}  
In the visual question answering (VQA) task, each input $I = (\text{image}, \text{question})$ is paired with a ground-truth answer $G$. The retriever’s goal is to identify neighbors that help the LMM generate the correct answer when used as in-context demonstrations. Similar to classification, we evaluate the model’s prediction without any neighbors, $P_0 = \text{LMM}(I)$, and with a candidate neighbor $N$, $P_N = \text{LMM}(N, I)$. Since the output space in VQA is discrete (e.g., multiple-choice or short text answers), correctness can be determined by exact matching against $G$. A neighbor is therefore labeled as \emph{positive} if its inclusion leads to the correct answer, and \emph{negative} otherwise:

\begin{equation}
    s(N, I) =
    \begin{cases}
        1, & \text{if } P_N = G, \\
        0, & \text{otherwise}.
    \end{cases}
\end{equation}

This binary supervision enables the retriever to focus on neighbors that genuinely improve reasoning over the image–question pair, while filtering out semantically similar but unhelpful examples.

\paragraph{Training Objective.} Using the generated supervision, we train a retriever to learn an embedding space that aligns with in-context usefulness. Our objective is that positive examples should cluster near their queries, while hard negatives are pushed farther apart. To achieve this, we adopt a supervised contrastive learning framework~\cite{khosla2020supervised}. For each query sample $x_i$, the training mini-batch includes the query itself, its positive set $\mathcal{P}_{x_i}$, and its negative set $\mathcal{N}_{x_i}$. The retriever is optimized with the supervised contrastive loss

\begin{equation}
   \mathcal{L}_{i,j} = -\log\frac{\exp\left(\text{sim}\left({z}_{i}, {z}_{j}\right)/\tau\right)}{\sum^{B}_{k=1}{1}_{[k\neq{i}]}\exp\left(\text{sim}\left({z}_{i}, {z}_{k}\right)/\tau\right)},
\end{equation}
where ${z}_{i}$ denotes the feature representation of $x_i$, $\tau$ is a temperature parameter, and $\text{sim}(u, v) = u^{T}v / (\|u\|\|v\|)$ is the cosine similarity. This formulation encourages the retriever to assign higher similarity to query–positive pairs and lower similarity to query–negative pairs. Over time, the learned feature space evolves to reflect the true in-context effectiveness of examples rather than their superficial similarity.

Once trained, GRIP replaces standard similarity-based retrieval at inference time. Given a new query, the retriever efficiently ranks candidate examples in the learned embedding space and selects those most likely to improve the model’s prediction. This enables the system to construct prompts adaptively, tailoring them to the model’s actual behavior and closing the gap between what appears semantically related and what is genuinely useful for multimodal in-context learning.

\paragraph{Retriever Architecture.} Our retriever builds upon a pre-trained CLIP vision encoder, specifically the ViT-L/14 model trained on the LAION-2B English subset of LAION-5B (\cite{schuhmann2022laion}). We fine-tune this encoder within a supervised contrastive learning framework to align its embedding space with in-context usefulness. The implementation extends the publicly available supervised contrastive learning codebase from SupContrast (\cite{khosla2020supervised}), adapting it for our feedback-guided retriever training.

\section{Experiments}

\paragraph{Tasks and Datasets.}  
We evaluate our approach on three representative tasks: (1) visual question answering (VQA), (2) image captioning, and (3) image classification. For VQA, we use the ScienceQA (\cite{lu2022learn}) and SEED-Bench (\cite{li2023seed}) datasets; for image captioning, we use MS COCO (\cite{chen2015microsoft}), and for image classification, we employ the Describable Textures Dataset (DTD) (\cite{cimpoi14describing}), UC Merced Land Use Dataset (\cite{yang2010bag}), and Oxford-IIIT Pet Dataset (\cite{parkhi2012cats}). 

Our retriever operates purely on visual embeddings and uses no textual
input at retrieval time. We therefore compare it against vision-based
retrievers such as nearest-neighbor selection in ViT, DINO, and CLIP feature
spaces, so that every method draws on the same visual information and
any performance difference reflects the quality of retrieval rather than
access to an additional modality. This ensures a fair evaluation of retrieval performance under equivalent information constraints. For VQA and classification, we report accuracy, while for image captioning, we use the CIDEr score as the primary evaluation metric. 

\paragraph{Models.}  
We conduct experiments using two large multimodal models: Qwen2.5-VL-7B-Instruct (\cite{yang2025qwen2}) and Idefics2-8B (\cite{laurenccon2024matters}). These models were chosen for their ability to process query and in-context images jointly within the input context. 

Our retriever is initialized from the CLIP ViT-L/14 encoder (\cite{radford2021learning}), pre-trained on the LAION-2B English subset of LAION-5B (\cite{schuhmann2022laion}) using OpenCLIP (\cite{ilharco_gabriel_2021_5143773}). We fine-tune this encoder for 200 epochs using stochastic gradient descent (SGD) with an initial learning rate of 0.005, which is decayed following a cosine annealing schedule.

\begin{table}
  \centering
\caption{Performance comparison of our feedback-guided retriever, GRIP, against zero-shot inference and similarity-based retrieval baselines across three tasks: image classification (DTD, UC Merced, Oxford Pets), visual question answering (ScienceVQA, SEED-Bench), and image captioning (COCO). For classification and VQA we report accuracy, while for image captioning, we use the CIDEr score. Best results in bold.
}
  \vspace{0.2cm}
  \label{tab:results-main}
  \begin{tabular}{@{}cccccccc@{}}
    \toprule
    \multirow{3}{*}{Model} & \multirow{3}{*}{Method} & \multicolumn{3}{c}{Classification} & \multicolumn{2}{c}{VQA} & Captioning \\
    \cmidrule(r){3-5}\cmidrule(r){6-7} \cmidrule{8-8}
    & & DTD & UC Merced & Pets & ScienceVQA & SEED  & COCO \\  
    \midrule
    \multirow{5}{*}{Qwen2.5-VL-7B} & Zero Shot & 48.7 & 69.0 & 75.9 & 78.2 & 69.5 & 62.3 \\
    & ViT & 52.2 & 71.4 & 79.0 & 81.0 & 70.2 & 71.7 \\
    & DINO & 52.3 & 70.5 & 78.1 & 81.9 & 69.8 & 72.0 \\
    & CLIP & 51.5 & 70.0 & 78.2 & 81.6 & 69.2 & 69.0\\
    & \textbf{GRIP (Ours)} &  \textbf{53.0} & \textbf{74.5} & \textbf{83.9}  & \textbf{84.2} & \textbf{71.4} & \textbf{72.5} \\
    \midrule
    \multirow{5}{*} {Idefics2-8B} & Zero Shot & 43.7 & 66.4 & 55.0 & 79.2 & 66.3 & 78.1 \\
    & ViT & 54.8 & 84.2 & 88.1 & \textbf{84.8} & 67.4 & \textbf{80.9} \\
    & DINO & 54.1 & 85.7 & 86.2 & 84.6 & 65.4 & 79.8\\
    & CLIP & 55.8 & \textbf{85.9} & 86.2 & 83.1 & 66.6 & 80.7 \\
    & \textbf{GRIP (Ours)} & \textbf{61.3} & 85.2 & \textbf{88.5} & 84.7 & \textbf{68.2} & 79.9 \\
    \bottomrule
  \end{tabular}
\end{table}

\subsection{Our Experiments on Multimodal In-Context Learning}
We compare our proposed GRIP retriever against several baselines to evaluate its effectiveness in improving multimodal in-context learning (M-ICL). As a lower bound, we include a \emph{zero-shot} baseline, where the model receives no in-context examples. In addition, we evaluate three standard similarity-based retrievers that select the most visually similar samples in pre-trained feature spaces derived from ViT, DINO, and CLIP. Since our retriever operates purely on visual embeddings, without incorporating textual information, we restrict our comparisons to vision-based methods to ensure fairness and consistency.

Table~\ref{tab:results-main} summarizes the results across the three tasks: image classification, visual question answering (VQA), and image captioning. Overall, we observe that similarity-based retrievers already outperform the zero-shot baseline, confirming that providing relevant visual examples benefits M-ICL performance. GRIP improves further, it achieves the best result on all six tasks for Qwen2.5-VL-7B, and on three of six tasks for Idefics2-8B, remaining within one point of the strongest baseline in the other cases, demonstrating that learning from feedback enables the retriever to select examples that are genuinely beneficial for the model.

For image classification, GRIP beats the best similarity baseline on five of the six dataset settings. The strongest improvements are on DTD with Idefics2-8B ($+5.5$ points) and Oxford Pets with Qwen2.5-VL-7B ($+4.9$ points). GRIP also outperforms the best baseline on the remaining Qwen2.5-VL-7B datasets and on Pets for Idefics2-8B, trailing only on UC Merced with Idefics2-8B by 0.7 points.  When compared to the zero-shot baseline, the improvements are substantial, reaching $+33.5$ points (Oxford Pets, Idefics2-8B).

For VQA, GRIP achieves the best accuracy on three of the four VQA settings.
On Qwen2.5-VL-7B it improves over the best similarity baseline by $+2.3$ points on ScienceVQA and by $+1.2$ points on SEED-Bench. On Idefics2-8B, GRIP improves on SEED-Bench by $+0.8$ and is on par on ScienceVQA. This demonstrates that feedback-guided retrieval can better capture contextual relevance beyond mere visual similarity, improving reasoning-based tasks that rely on multimodal grounding.  

For image captioning, GRIP yields the best CIDEr score (72.5) on Qwen2.5-VL-7B, about 10 points above zero-shot. On Idefics2-8B, where zero-shot captioning is already strong (78.1), GRIP is competitive with the similarity retrievers. These gains indicate that even for open-ended generation tasks, retrieval tuned through model feedback can enhance the descriptive quality of outputs.  

In summary, these results demonstrate that GRIP improves consistently over similarity-based retrieval on Qwen2.5-VL-7B and is best-or-competitive on Idefics2-8B, with the best gains on classification. These results support the value of explicitly aligning the retrieval space with the model's in-context performance, using model feedback rather than relying on visual proximity alone.

\begin{table}
  \centering
    \caption{Cross-model generalization of GRIP. We evaluate the ability of our feedback-guided retriever to transfer across multimodal models. We train GRIP using guidance from one large multimodal model (either Qwen2.5-VL-7B or Idefics2-8B) and evaluate it on the other. The table reports the performance gap relative to the retriever trained and evaluated on the same model (see corresponding results in Table~\ref{tab:results-main}). Improvements over the original model-specific retriever are shown in \textcolor{green}{green}, while decreases are shown in \textcolor{red}{red}. Performance of the zero-shot approach and CLIP are the same as in Table~\ref{tab:results-main}. We report accuracy. Best results in bold.}
    \vspace{0.2cm}
  \begin{tabular}{@{}cccccc@{}}
    \toprule
    \multirow{3}{*}{Model} & \multirow{3}{*}{Method} & \multicolumn{2}{c}{Classification} & \multicolumn{2}{c}{VQA} \\
    \cmidrule(r){3-4}\cmidrule(r){5-6} 
    & & UC Merced & Pets & ScienceVQA & SEED \\
    \midrule
    \multirow{3}{*}{Qwen2.5-VL-7B} & Zero Shot & 69.0 & 75.9 & 78.2 & 69.5 \\
    & CLIP Similarity Based & 70.0 & 78.2 & 81.6 & 69.2 \\
    & \textbf{GRIP} (Idefics2-8B) & \textbf{76.6} (\textcolor{green}{+2.1}) & \textbf{79.3} (\textcolor{red}{-4.6}) & \textbf{84.7} (\textcolor{green}{+0.5}) & \textbf{71.4} (\textcolor{green}{+0.0}) \\
    \midrule
    \multirow{3}{*}{Idefics2-8B} & Zero Shot & 66.4 & 55.0 & 79.2 & 66.3 \\
    & CLIP Similarity Based & \textbf{85.9} & 86.2 & 83.1 & 66.6 \\
    & \textbf{GRIP} (Qwen2.5-VL-7B) & 80.0 (\textcolor{red}{-5.2}) & \textbf{88.1} (\textcolor{red}{-0.4}) & \textbf{84.6} (\textcolor{red}{-0.1}) & \textbf{68.6} (\textcolor{green}{+0.4}) \\
    \bottomrule
  \end{tabular}
  \label{tab:results-extensibitly}
\end{table}

\subsection{Our Experiments on Cross-Model Retrieval}
An important question in feedback-guided retrieval is whether a retriever trained with feedback from one LMM can generalize to another, potentially reducing the need for retraining and lowering computational cost. To investigate this, we evaluate our retriever across models by training it with feedback from one LMM and applying it directly to another. Specifically, we train GRIP on Idefics2-8B and test it on Qwen2.5-VL-7B, and vice versa, covering both image classification and visual question answering (VQA) tasks. 

The results, summarized in Table~\ref{tab:results-extensibitly}, demonstrate that cross-model retrieval remains highly effective and, in several cases, even yields performance gains. When trained on Idefics2 and evaluated on Qwen2.5-VL, GRIP achieves improvements over both the zero-shot baseline and the CLIP-based similarity retriever across most datasets. For instance, on the UC Merced dataset, cross-model GRIP improves by 2.1 points relative to the model-specific retriever, and it also shows small but consistent gains on ScienceVQA and SEED. Conversely, when trained on Qwen2.5-VL and evaluated on Idefics2, GRIP maintains strong performance, with only minor drops and even slight improvements on SEED. 

These results highlight two key observations. First, retrievers can transfer well to other models with minimal loss, suggesting that feedback-based supervision captures model-agnostic patterns of in-context usefulness. Second, the robustness of cross-model GRIP indicates that feedback-guided retrieval can serve as a portable component across LMMs, eliminating the need to retrain the retriever for every new target model. This cross-model consistency demonstrates the practicality and scalability of our approach for large-scale deployment in diverse multimodal systems.

\begin{table}
  \centering
  \caption{Evaluation on proprietary multimodal models. We test GRIP retrievers trained on open-source models (Qwen2.5-VL-7B and Idefics2-8B) on closed-source LMMs without any retraining. Both variants of GRIP consistently outperform the zero-shot baseline and the CLIP similarity-based retriever, demonstrating strong cross-model generalization. We report accuracy. Best results in bold.}
  \vspace{0.2cm}
  \begin{tabular}{@{}cccc@{}}
    \toprule
    \multirow{3}{*}{Method} & GPT 4o & \multicolumn{2}{c}{Gemini 2.5 Flash Lite}\\
    \cmidrule(r){2-2}\cmidrule(r){3-4}
    & ScienceVQA & ScienceVQA & SEED \\
    \midrule
    Zero Shot & 79.3 & 67.1 & 40.7 \\
    CLIP Similarity Based & 84.8 & 75.1 & 43.5 \\
    \textbf{GRIP} (Qwen2.5-VL-7B) & 85.0 & \textbf{75.9} & 43.7 \\
    \textbf{GRIP} (Idefics2-8B) & \textbf{85.9} & 75.2 & \textbf{44.2} \\
    \bottomrule
  \end{tabular}
  \label{tab:results-gpt}
\end{table}

\subsection{Our Experiments on Retrieval for Proprietary Models}
Testing feedback-guided retrieval on proprietary large multimodal models (LMMs) such as GPT-4o and Gemini provides a strong test of generalization and scalability. Since these models are accessible only via APIs, directly training a feedback-based retriever on them is impractical due to both computational and financial costs. To address this challenge, we investigate whether retrievers trained on publicly available open-source models can be applied to closed-source systems without additional fine-tuning.

Table~\ref{tab:results-gpt} summarizes the results on the ScienceVQA and SEED benchmarks for GPT-4o and Gemini 2.5 Flash Lite. We evaluate our approach on Gemini 2.5 Flash Lite rather than the original Gemini model, as the latter is significantly more resource-intensive in terms of both inference time and API cost. Moreover, demonstrating consistent improvements on this constrained version provides a stronger indication of the empirical usability and robustness of our retrieval strategy under realistic deployment conditions. We observe that both GRIP variants outperform the zero-shot baseline and the
CLIP similarity retriever on all datasets, despite never being trained on the target model. For GPT-4o, GRIP trained with Idefics2-8B feedback achieves the best accuracy on ScienceVQA (85.9\%), $+6.6$ points over zero-shot and $+1.1$ points over CLIP. For Gemini 2.5 Flash Lite, GRIP trained with Qwen2.5-VL-7B feedback attains the highest ScienceVQA accuracy
(75.9\%, $+0.8$ over CLIP), and the GRIP variants give the best SEED-Bench
result (44.2\% with Idefics2-8B feedback, $+0.7$ over CLIP). 

These findings confirm that feedback-guided retrievers trained on open multimodal models can generalize effectively to closed-source, API-based systems, enhancing their in-context learning capabilities without model-specific retraining. This property is particularly valuable for practical deployment, as it enables cost-efficient adaptation of high-performing retrievers to powerful but inaccessible proprietary LMMs. By using feedback from open-source models like Qwen2.5-VL and Idefics2, our approach offers a scalable way to improve the reasoning and contextual understanding of advanced multimodal systems.

\subsection{Study on the choice of a backbone for GRIP}

To analyze the influence of the underlying visual encoder on our feedback-guided retriever, we conduct a study comparing different backbone architectures within GRIP. Specifically, we evaluate two variants of GRIP: one using a ViT backbone and another using a CLIP-based backbone, both trained with the same feedback-guided supervision. We assess their performance on image classification and visual question answering (VQA) tasks using both Qwen2.5-VL-7B and Idefics2-8B as the downstream multimodal models.

The results in Table~\ref{tab:ret-model} show that GRIP performs strongly with either visual backbone, with the CLIP-based encoder giving the best results across both downstream models. Notably, the difference between the two backbones is small relative to GRIP's gains over similarity-based retrieval (Table~\ref{tab:results-main}), indicating that the main driver of performance is the feedback-guided learning process at the core of our approach, while the backbone provides a complementary improvement. GRIP is therefore not tied to a particular encoder: it delivers strong retrieval with both ViT and CLIP, and CLIP's image–text pretraining adds a further edge, which leads us to adopt it as the default in our main experiments.

\begin{table}[t]
  \centering
  \caption{Comparison of GRIP using different visual encoders (ViT vs.\ CLIP) as backbones across image classification and VQA tasks for Qwen2.5-VL-7B and Idefics2-8B. We report accuracy.}
    \vspace{0.2cm}
  \begin{tabular}{@{}cccccc@{}}
    \toprule
    \multirow{3}{*}{Model} & \multirow{3}{*}{Method} & \multicolumn{2}{c}{Classification} & \multicolumn{2}{c}{VQA} \\
    \cmidrule(r){3-4}\cmidrule(r){5-6}
    & & DTD & UC Merced & ScienceVQA & SEED \\  
    \midrule
    \multirow{2}{*}{Qwen2.5-VL-7B} & GRIP w/ ViT  & 53.6 & 72.3 & 82.5 & 70.8 \\
    & GRIP w/ CLIP & 53.0 & 74.5 & 84.2 & 71.4 \\
    \midrule
    \multirow{2}{*} {Idefics2-8B} & GRIP w/ ViT  & 58.7 & 86.1 & 84.4 & 67.3 \\
    & GRIP w/ CLIP & 61.3 & 85.2 & 84.7 & 68.2 \\
    \bottomrule
  \end{tabular}
  \label{tab:ret-model}
\end{table}

\section{Related Work}

\paragraph{Multimodal In-Context Learning and Demonstration Selection.}
Multimodal In-Context Learning (M-ICL) enables large multimodal models (LMMs) to adapt to new tasks by conditioning on a small set of image--text demonstrations at inference time. M-ICL arises from the training objectives of large vision-language models, such as Flamingo \citep{alayrac2022flamingo}, which are trained on sequences of interleaved images and text. Since the choice of demonstrations strongly affects performance, several works study example selection for M-ICL. RICES \citep{yang2022empirical} retrieves demonstrations using visual similarity, while MMICES \citep{chen2025can} and MUIER \citep{luo2024does} incorporate both visual and textual signals for multimodal example selection. More recently, CIRCLES \citep{xiong2026retrieving} moves beyond passive nearest-neighbor retrieval by constructing counterfactual-style demonstration sets through attribute-guided composed image retrieval. GRIP complements these directions by taking a model-centric view of demonstration selection, it learns which examples are beneficial from the LMM's own observed behavior.

\paragraph{Learning-Based and Feedback-Guided Retrieval.}
Beyond heuristic retrieval, learning-based demonstration selection has also been extensively studied for in-context learning. Prior work shows that retrieval optimized for downstream performance can outperform nearest-neighbor selection \citep{rubin-etal-2022-learning, luo2023dr}. Other methods use model feedback as supervision, for example, LLM-R \citep{wang2023learning} distills utility-based reward signals into a bi-encoder retriever, Syntriever \citep{kim2025syntriever} aligns retrieval with black-box LLM preferences using synthetic supervision, and RDES \citep{wang2024demonstration} formulates demonstration selection as a reinforcement-learning problem. These methods establish the value of feedback-driven retrieval, but they primarily focus on text-only in-context learning. GRIP brings this principle to multimodal ICL in a simple form: it converts task-level feedback from LMM predictions into supervision for a compact vision-only retriever, aligning the retrieval space with actual in-context utility.

\section{Limitations and Future Work}

While our work advances the understanding of feedback-guided retrieval in multimodal in-context learning, it leaves open a number of exciting directions for future research.

Our current approach focuses exclusively on visual information when learning and retrieving examples. Although this design choice simplifies the analysis and isolates the visual contribution to retrieval quality, multimodal systems naturally benefit from joint visual-textual reasoning. Incorporating textual information, such as captions, questions, or answers, into the retrieval process can provide richer contextual cues and further enhance the alignment between the retrieved examples and the target query. Extending GRIP to operate in a fully multimodal feature space thus represents an important next step.

Our framework currently considers a single in-context example per query. While this allows for clearer attribution of individual example influence, it does not capture the cumulative or potentially synergistic effects that can arise when multiple examples are presented jointly. Future work will explore how to retrieve, combine, and order multiple in-context examples, and how large multimodal models interpret such aggregated context. This includes investigating adaptive selection strategies that dynamically determine the number and arrangement of retrieved samples based on task characteristics or model feedback.

Overall, we view our work as a step toward developing adaptive, feedback-driven retrieval systems that more effectively exploit the multimodal capabilities of next-generation foundation models.

\section{Conclusion}
In this work, we revisited the foundations of multimodal in-context learning (M-ICL) and questioned the common assumption that semantically or visually similar examples always provide the most useful context for LMMs. Through a systematic analysis, we showed that similarity-based retrieval does not consistently yield optimal in-context performance. This finding highlights a fundamental gap between visual similarity and contextual usefulness in M-ICL, motivating the need for retrieval strategies that explicitly consider model feedback.

To address this, we introduce a vision-only learned retrieval framework, called GRIP, that leverages feedback from LMMs to identify examples that truly enhance model reasoning and generalization. Unlike conventional similarity-driven or heuristic-based retrievers, GRIP learns to differentiate between beneficial and detrimental in-context examples through contrastive training. Across a range of tasks, including image classification, captioning, and visual question answering, GRIP improved consistently over similarity-based retrieval on Qwen2.5-VL-7B and remained best-or-competitive on Idefics2-8B, with its largest gains in classification, demonstrating its ability to select examples that lead to more effective multimodal adaptation.

Finally, we demonstrated that retrievers trained with feedback from one model can generalize to others, including closed-source API-based systems such as GPT-4o and Gemini. This cross-model transferability shows that feedback-guided retrieval can enhance the in-context learning capabilities of powerful proprietary models without the need for retraining or costly fine-tuning, offering a scalable and practical path for improving multimodal reasoning at scale.

Overall, our study not only challenges the conventional similarity paradigm but also establishes feedback-guided retrieval as a principled and effective direction for advancing multimodal in-context learning in both open and closed model ecosystems.

\bibliography{main}
\bibliographystyle{tmlr}


\end{document}